\documentclass{article}

\usepackage{graphicx} 
\usepackage{textcomp} 
\usepackage{gensymb} 
\usepackage{indentfirst} 
\usepackage{enumerate} 
\usepackage{authblk} 
\usepackage[style=ieee, backend=biber]{biblatex} 
\usepackage{amsmath, amssymb, amsfonts} 
\usepackage{mathtools} 
\usepackage{xspace} 
\usepackage{makecell} 
\usepackage{tabularx} 
\usepackage{float} 
\usepackage[x11names]{xcolor} 
\usepackage{subcaption} 
\usepackage[colorlinks=true, linkcolor=blue, citecolor=blue, urlcolor=blue]{hyperref} 

\providecommand{\keywords}[1]
{
  \small	
  \textbf{\textit{Keywords---}} #1
}

\title{From Images to Detection: Machine Learning for Blood Pattern Classification}
\author[1]{Yilin Li \thanks {ilnli@ucdavis.edu}}
\author[1]{Weining Shen \thanks{weinings@uci.edu}}
\affil[1]{University of California, Davis}
\affil[1]{University of California, Irvine}

\date{}

\begin{document}

\maketitle

\newpage

\begin{abstract}
    Bloodstain Pattern Analysis (BPA) helps us understand how bloodstains form, with a focus on their size, shape, and distribution. This aids in crime scene reconstruction and provides insight into victim positions and crime investigation. One challenge in BPA is distinguishing between different types of bloodstains, such as those from firearms, impacts, or other mechanisms. Our study focuses on differentiating impact spatter bloodstain patterns from gunshot bloodstain patterns. We distinguish patterns by extracting well-designed individual stain features, applying effective data consolidation methods, and selecting boosting classifiers. As a result, we have developed a model that excels in both accuracy and efficiency. In addition, we use outside data sources from previous studies to discuss the challenges and future directions for BPA.
\end{abstract}

\keywords{Bloodstain pattern analysis, Forensic statistics, XGBoost, Random forest, Feature extraction, Image processing}

\section{Introduction}

Bloodstains are among the most prevalent and vital evidence encountered in violent crime investigations, making Bloodstain Pattern Analysis (BPA) a key component in forensic science \cite{wonder2015bloodstain}. BPA offers insights into various investigative aspects in crimes, such as determining the area of convergence and the point of origin of bloodstains \cite{de2011improving}\cite{varney2011locating}\cite{bevel1002ross}\cite{knock2007predicting}, identifying the generating mechanism's type and direction, and inferring victim positions or movements. These insights play an important role in forensic investigations by addressing key questions such as the nature of the event, the location of the crime, and the chronological sequence of events \cite{james2005principles}. Unlike techniques based on chemical composition, BPA focuses on analyzing the size, shape, and distribution of bloodstains to extract useful information \cite{james2005principles}.

One of the major challenges in  BPA is distinguishing between different types of bloodstain patterns, particularly determining whether a pattern results from firearms, impacts, or other mechanisms, and whether it is an impact spatter or a cast-off spatter. Early classification tasks are conducted by pattern analysts, which tends to result in subjective conclusions based on the knowledge background of the experts \cite{laber2014reliability}\cite{zou2022towards}, and it can easily be influenced by contextual bias \cite{macdonell1971flight}. A black box study reveals frequent errors and contradictions among BPA analysts, which can pose serious consequences in casework and court testimony \cite{hicklin2021accuracy}. These uncertainties call for more objective and scientific methods \cite{national2009strengthening}. 
Classical statistical tools, such as the likelihood ratio test \cite{zou2022towards}\cite{meuwly2017guideline}\cite{neumann2006computation}\cite{bolck2009different}, can be used as classifiers for various pattern types. Machine learning methods have also been used in forensic science for prediction and classification purposes, including age estimation based on DNA methylation \cite{aliferi2018dna}, differentiating individuals by chemical components of fingerprints \cite{gorka2023differentiating}, and identifying forged handwritten signatures \cite{gaborini2017towards}. Commonly used methods include SVM (Supportive Vector Machine), LDA (Linear discriminant analysis), QDA (Quadratic discriminant analysis) \cite{arthur2018automated}, and random forests \cite{liu2020automatic}.

In this paper, we focus on the classification of mechanisms generating blood spatter patterns, to be specific, gun-shot backspatters and impact beating spatters. These patterns occur when an object strikes a blood source, propelling drops into the air before they land on a solid surface referred to as the target. The dataset studied in this article is publicly accessible, consisting of 169 blood spatter patterns, which are a subset of bloodstain patterns, with 68 gunshot backspatters and 61 blood impact beating spatters. Researchers have conducted studies on them to formulate classifiers with the likelihood ratio test \cite{zou2022towards} and the random forest \cite{liu2020automatic}. The former leverages directional statistics to extract interpretable angular features but results in low classification accuracy. In contrast, the random forest approach \cite{liu2020automatic} yields higher precision but is based on tedious features, involving complex curve approximation and derivative computation.  


The contribution of our paper is three-fold. First, we introduce a new feature construction methodology that simplifies the process of converting bloodstain images into numerical data. Unlike previous studies that relied on complex and less interpretable features or suffered from unsatisfactory classification performance, our method considers basic local features and tone-related features that are both interpretable and produce comparable classification accuracy. Second, in addition to using a random forest model, we apply XGBoost, a boosting algorithm that has not been used in BPA research, and demonstrate its advantages, such as efficiently handling missing data and faster computation. Third, we develop a new metric for assessing feature importance in classification models, called the \textit{stability importance score} (SIS). Unlike traditional importance metrics in bagging or boosting models that only reflect relative importance and cannot be averaged across different model fits, SIS measures the importance by tracking how often a feature ranks among the top 10 most significant across multiple model runs. This metric provides a more consistent and interpretable measure of feature importance. 


The rest of the paper is organized as follows. We introduce the data set and its preprocessing in Section \ref{sec:data}. We then describe feature construction in Section \ref{sec: feature_design}. In Section \ref{sec4}, we present the results of XGBoost and other classification approaches. We discuss the challenges associated with data quality and future directions in Section \ref{sec: discussion}.


\section{Data}
\label{sec:data}

\subsection{Data Source}
The dataset used in this study is open source \cite{attinger2019data}\cite{attinger2018data}, consisting of 169 blood spatter patterns, which are a subset of bloodstain patterns. Among these patterns, there are 68 gunshot backspatters and 61 blood impact beating spatters. Impact spatters are created when blood is atomized as a result of an impulsive force applied to the blood source, resulting in a unique pattern. On the other hand, gunshot backspatters are formed when blood is atomized by a bullet, moving in the opposite direction to the bullet's path, forming a distinct spatter pattern.
    \begin{figure}[H]
        \includegraphics[width = \textwidth]{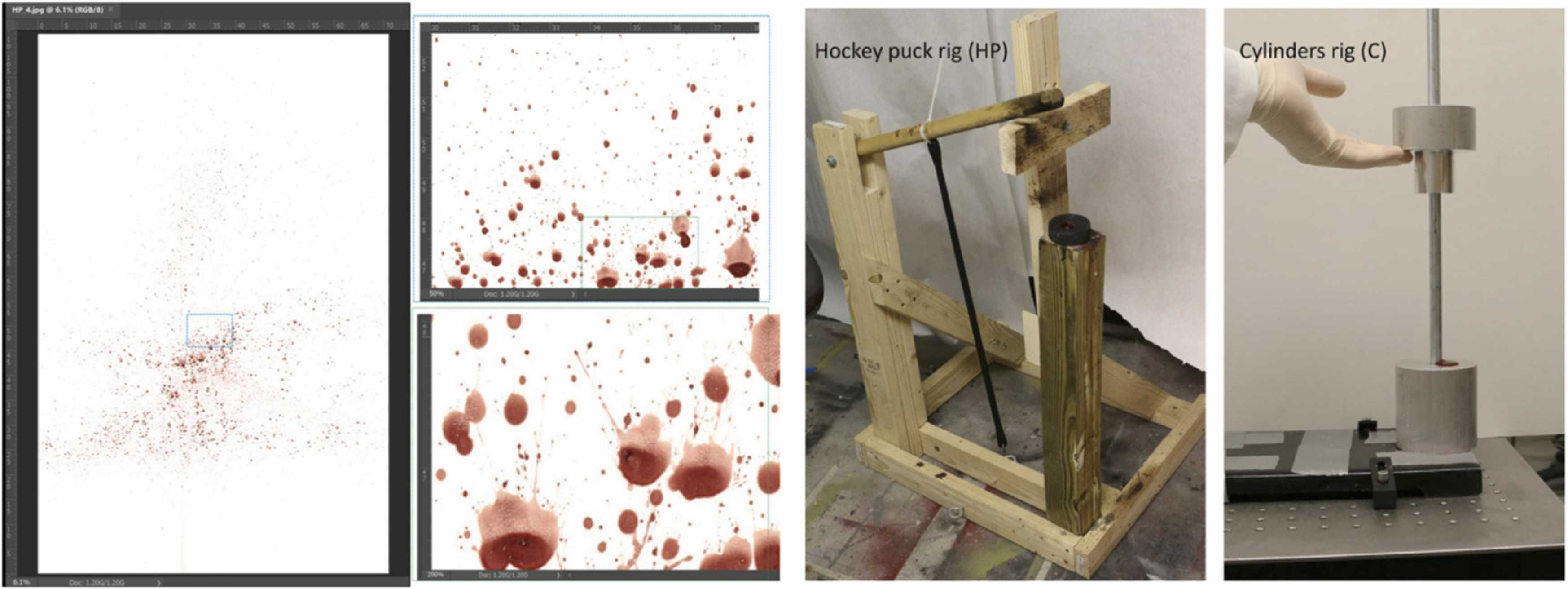}\caption{On the right is a sample of impact spatter pattern; the scale is on the edge. The test rigs used for generating the patterns is shown on the left. The figures are taken from \cite{attinger2018data}.}
        \label{figure:impact_example}
    \end{figure}

During the experiments, most blood spatters were generated on flat poster board sheets, typically using the smoother side. Some experiments involved the exploration of the rougher side or the use of butcher paper. Larger targets, up to $136 \times 110$ $cm^2$ in size, were created by combining two or four poster board sheets. Fresh swine blood containing anticoagulants was employed to create 94 impact and backspatter patterns within an indoor environment. These patterns were allowed to air dry naturally before scanning at a high resolution of 600 dpi. To create impact spatter patterns, two different setups were employed. The first involved rapid compression of blood between a cylindrical wooden dowel and a flat surface.  In the second setup, blood was compressed between two flat surfaces using cylinders. Figure \ref{figure:impact_example} shows an example of an impact spatter pattern on different scales, and the rigs to generate these patterns \cite{zou2022towards}\cite{attinger2018data}.

\begin{figure}[H]
        \includegraphics[width = \textwidth]{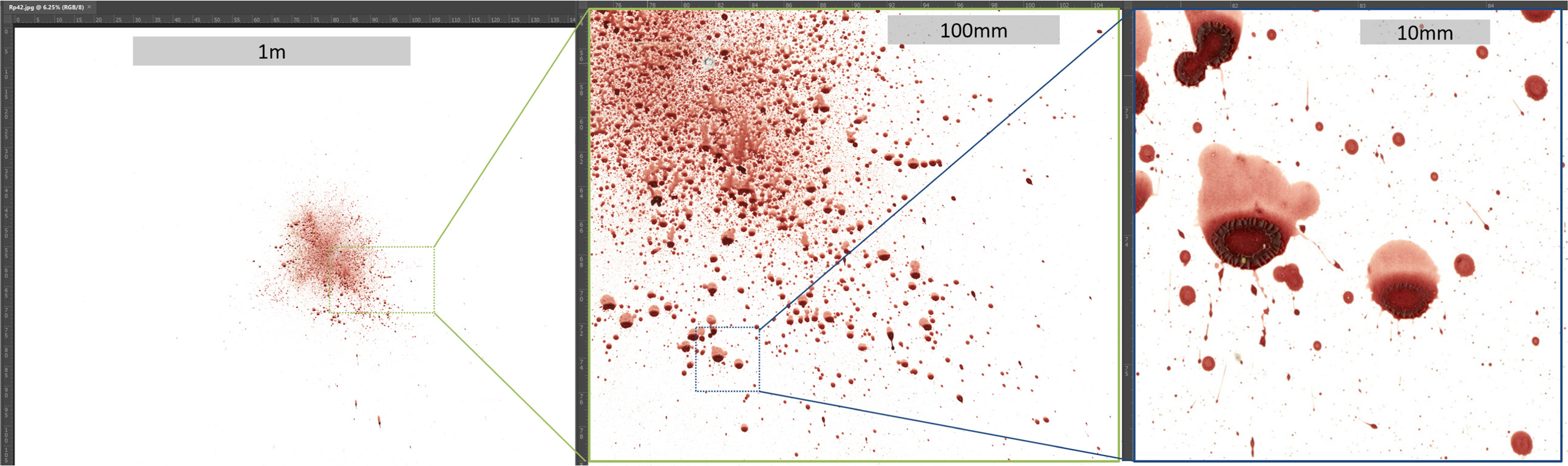}\caption{A gun shot pattern sample. The figures are taken from \cite{attinger2019data}.}
        \label{figure:gun_example}
    \end{figure}

To create gunshot backspatters, the blood for spatter formation was obtained from a foam or sponge soaked with blood or from a closed blood-filled cavity reservoir. In the latter case, a portion of paper was removed from one side of a foam board, the resulting cavity was filled with blood, and then sealed with clear packaging tape. This method showed greater reproducibility compared to the use of soaked foam or sponge. A cardstock target was placed vertically between the gun muzzle and the target to collect backward spattered drops, with the bullet trajectory running parallel to the ground. A sample is shown in Figure \ref{figure:gun_example}. 

From Figures \ref{figure:impact_example} and \ref{figure:gun_example}, we noticed that the bloodstain patterns were composed of a white background and red bloodstains in nearly elliptical shapes. Therefore, we approximated each bloodstain in the pattern with an ellipse and used the graphical properties of fitted ellipses to build features for the pattern \cite{zou2022towards}\cite{arthur2018automated}\cite{liu2020automatic}\cite{arthur2017image}. Subsequently, we employed classifiers trained on these features to differentiate between impact spatter patterns and gunshot patterns.

\subsection{Image Processing}
\begin{figure}[H]
    \centering
    \subfloat[\centering Original pattern]{\includegraphics[width=2.75cm]{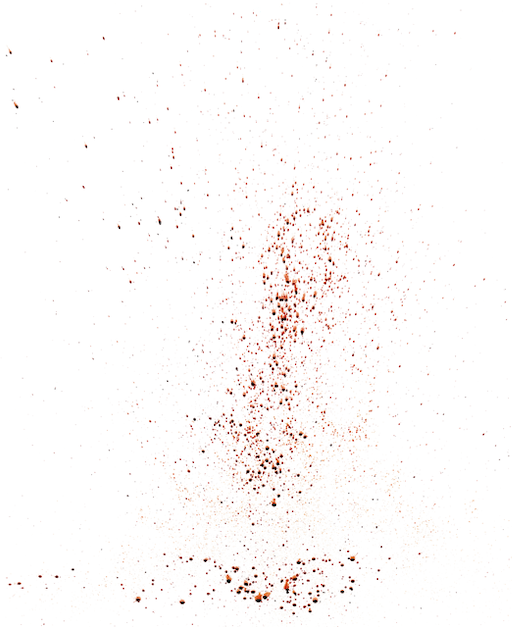} }%
    \qquad
    \subfloat[\centering After gray-scale transformation]{\includegraphics[width=2.75cm]{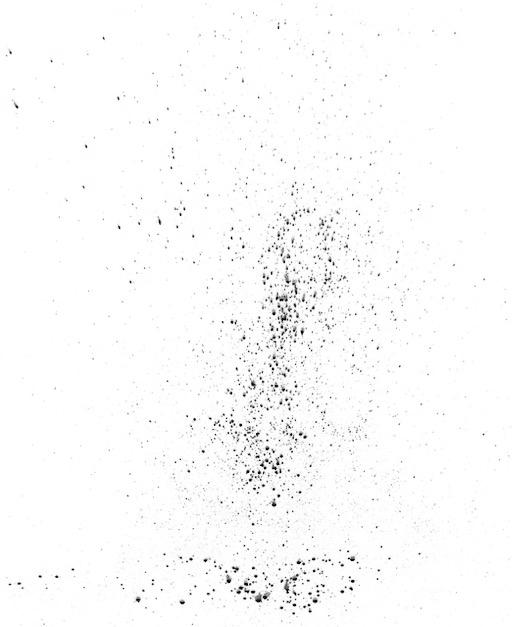} }%
    \qquad
    \subfloat[\centering After binarization and inversion]{\includegraphics[width=2.75cm]{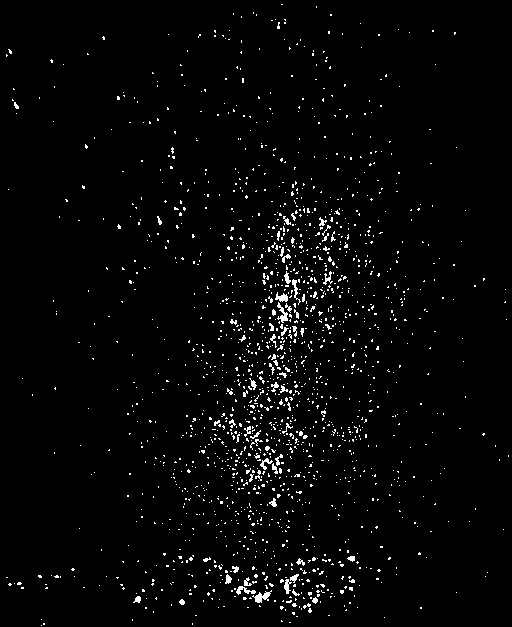} }%
    \caption{An image undergoing processing steps.}%
    \label{fig:Example_impact}%

\end{figure}
In MATLAB R2023a \cite{MATLAB}, we imported JPEG images and preprocessed them following the methods described in \cite{arthur2017image}. Initially, we converted the images to grayscale. While MATLAB's image processing functions primarily operate on pixels with higher values, in our case, the darker pixels representing stain locations were close to the value $0$, while pixels representing the white background had a value of $255$ (the maximum value in grayscale images). Therefore, we transformed the pixel values, $x$, to $255-x$, to obtain the inverse of the images. Then the grayscale images underwent binarization, where each pixel was assigned either a $0$ or $1$ based on a specific threshold. An example illustrating these steps is shown in Figure \ref{fig:Example_impact}. By the 8-connectivity principle \cite{MathWorksPixel}, pixels were considered connected if their edges or corners were touched. Pixels with a value of $1$ formed regions composed of connected pixels.  To distinguish between different stain elements, we labeled these regions with values $1, 2, 3, \ldots$, and assigned these values to the pixels within each region, ensuring that the pixels within the same region shared the same value. The labeled regions, now separated, would then be approximated by ellipses.

\section{Feature Engineering}
\label{sec: feature_design}

\subsection{Stain Features}
\begin{figure}[H]
    \centering
    \subfloat[\centering Original pattern]{\includegraphics[width=4cm]{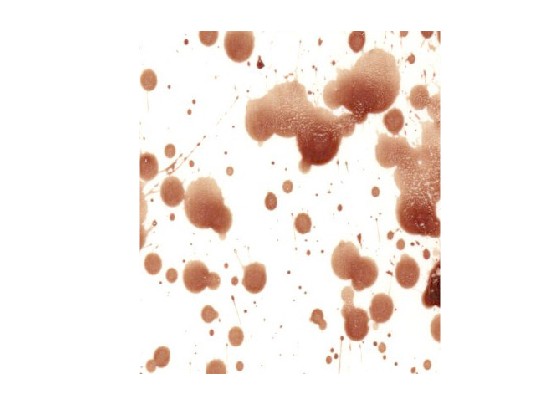} }%
    \qquad
    \subfloat[\centering After gray-scale transformation]{\includegraphics[width=4cm]{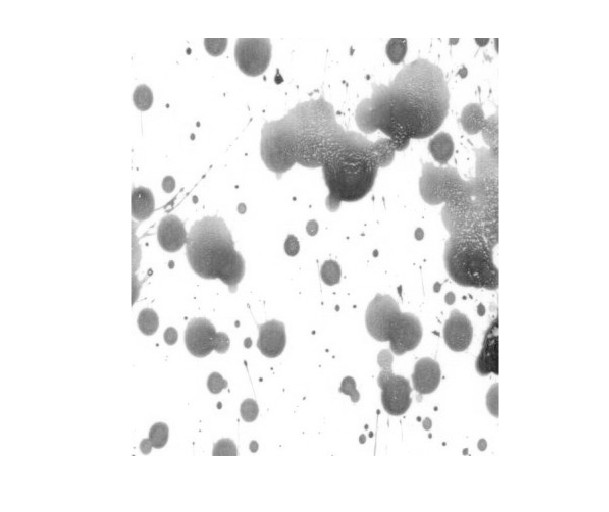} }%
    \qquad
    \subfloat[\centering After inversion and binarization]{\includegraphics[width=4cm]{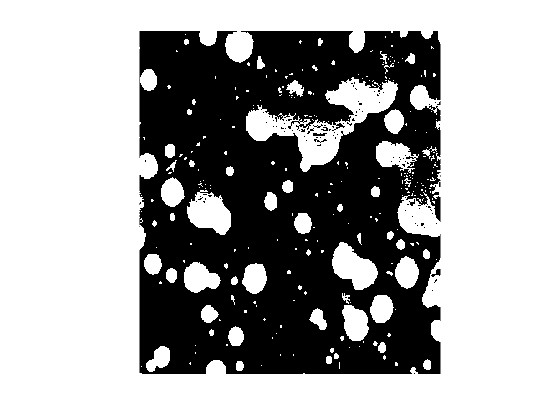} }%
    \qquad
    \subfloat[\centering After using \textit{regionprops} to detect and identify ellipses]{\includegraphics[width=4cm]{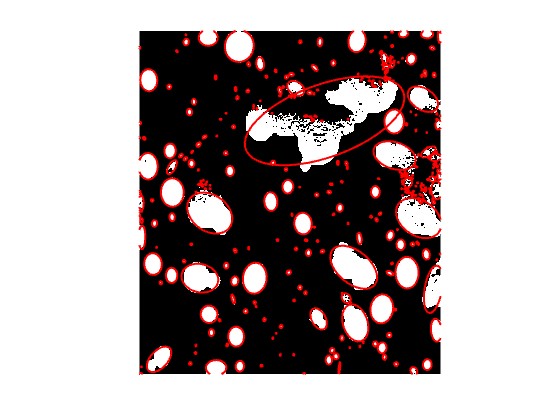} }%
    \caption{Example image undergoing processing and identification steps.}%
    \label{fig:Example_ellipse}%
\end{figure}

Using the built-in MATLAB function \textit{regionprops}, we are able to fit each connected labeled region in the data preprocessing steps with an ellipse, thus detecting and approximating each individual blood stain. Figure \ref{fig:Example_ellipse} provides a zoomed-in view of the pre-processing and fitting steps of one pattern, where non-overlapping stains are detected precisely. For overlapping stains, \cite{zou2022towards} applied the algorithm in \cite{zou2021recognition} to detect individual stains. However, we did not attempt to separate them, retaining the fitted result where several stains are identified as one element. Remarkably, the classifiers exhibit high accuracy despite the presence of overlapping elements. Addressing this issue could be considered for future research. After filtering out the images that do not meet certain criteria as outlined at the end of this section, we detect 150 to 95921 ellipses in one figure, with a mean value of 16725
With \textit{regionprops}(or with simple linear transformations), the following features for individual regions can be obtained. 

\begin{enumerate}
    \item Area (scalar): The actual number of pixels in the region. 
    \item Major/Minor axis length (scalar): The length (in pixels) of the major/minor axis of the ellipse that has the same normalized second central moments as the region.
    \item Vertical angle (scalar): This is defined as the $90\degree$ minus orientation, representing the angle between the x axis and the major axis of the ellipse that shares the same second moments as the region. Its value is expressed in degrees and ranges from -90 degrees to 90 degrees.
    \item Centroid (2-d vector): The center of mass of the region.
    \item Solidity (scalar): Proportion of the pixels in the convex hull that are also in the region.
\end{enumerate}

We aim to fully capture both the graphical characteristics and the distribution characteristics of the stains, such as the \textit{impact angle}, which provides detailed information about how blood hits the target to form the stain and is commonly used to infer the area of origin (AO) \cite{joris2014calculation}.  As a result, we also construct features by applying non-linear computation over the measurements obtained directly from \textit{regionprops} in addition to the aforementioned properties. The indirectly extracted stain features are defined as follows:

\begin{enumerate}
\setcounter{enumi}{5}
    \item Impact angle (scalar): the angle at which the blood drop hits the target and is estimated by the ratio of the {\it minor axis length} to the {\it major axis length} of the ellipse \cite{balthazard1939etude}.
    
    \item Adjusted impact angle (scalar): This feature is formulated to accommodate blood stains with long tails that were not removed during processing. Denoted by $\epsilon$, the adjusted angle is defined using the formula in \cite{liu2020automatic}:
    \begin{align*}
        \epsilon &= \frac{\text{Minor~Axis~Length}}{\text{Major~Axis~Length}_{adj}},\\
        \text{Major~Axis~Length}_{adj} &= \text{Major~Axis~Length}\times \frac{\text{Filled~Area}}{\text{Ellips~Area}},
    \end{align*}
    where `Ellips Area' is the number of pixels in the fitted ellipse, defined as the product of `Minor Axis Length', `Major Axis Length' and the scale $\frac{\pi}{4}$. Also, `Filled Area' is an attribute derived from \textit{regionprops}, which counts the number of pixels within a region after all enclosed holes are filled.

    \item Distance (vector): We first defined the centroid of the pattern as the median position of all stains. Specifically, if we had $n$ stains within a pattern, each with centroid positions represented as $(a_i, b_i)$ for $i = 1, 2, ..., n$, then the centroid of the pattern was calculated as $(\text{median}( \{a_i\}^n_{i = 1}), \text{median}(\{b_i\}^n_{i = 1}))$. Then we computed the Euclidean distance between each individual stain's centroid to the centroid of the entire spatter pattern \cite{liu2020automatic}.
\end{enumerate}

The shade of the stains is another aspect of interest. To tell the difference between impact spatters and gunshot backspatters, it is important to consider that the blood may be altered by the bullets, potentially changing the color of the stains. Additionally, in real crime scenarios, gunshots often cause deep wounds that penetrate internal organs or muscles, while impacts typically affect the body's surface.  Consequently, the splattered blood in these two cases may originate from different parts of the human body and could exhibit different colors.

The continuous grayscale image, serving as an intermediate product, can encode the color shade and be used to create a scalar-valued color feature. We introduce the following shade-related features:

\begin{enumerate}
\setcounter{enumi}{8}
    \item Shade (scalar): The mean grayscale value of pixels in one detected connected element.
    
    \item Evenness (scalar):
    The standard deviation of the pixels' grayscale value in one connected element.
    
\end{enumerate}

We set the following criteria to filter out unwanted fitted ellipses following \cite{arthur2017image} and \cite{liu2020automatic}: (1) \textit{Eccentricity} is less than or equal to 0.3, (2) \textit{Impact angle} is less than $\pi/18$, (3) \textit{Solidity} is less than 0.75, (4) the ratio of the circle with minor axis length being diameter over \textit{FilledArea} is larger to 1, and (5) the ratio of \textit{Area} over \textit{FilledArea} is less than 0.95. In particular, the last criterion is effective in removing obvious overlapping stains according to \cite{liu2020automatic}.

\subsection{Summarized Features for Patterns}
For each pattern, there are several fitted ellipses, each with several extracted features. Our next job is to create a list of descriptive statistics based on these features. To achieve this goal, we need to decide which aspects of the features we are interested in, and  compute summary features based on those aspects.

We first divide the image into several regions to account for the potential differences in characteristics between the stains near the center and those at the periphery \cite{siu2017quantitative}. Then we define the following features on the basis of the regions.  
    \begin{enumerate}
        \item Graph-wide: We consolidate data globally by aggregating the features of all individual bloodstains within a single spatter pattern. This process allows us to create a comprehensive representation that captures the collective characteristics of the entire spatter pattern.
        \item Annuli: The annuli are organized as concentric rings, all centered around the centroid of the spatter pattern. An annulus, which represents the area enclosed between two concentric circles, forms a geometric framework. Each annulus can be customized to have a fixed width, as specified by the user, or it can dynamically adjust based on the greatest distance between each elliptical element and the image's center. We structure the pattern segregation according to \cite{siu2017quantitative}, where the spatter pattern is segmented into 40 annuli, each with a 2.5 cm width, radiating outward from the pattern's centroid. For the adaptive version, we computed the Euclidean distance between each individual stain and the pattern centroid. Then we set the median distance divided by 20 as the adaptive width to form 40 annuli.
        \item Rectangular bins: Another way to define bins is by partitioning the spatter pattern region into equidistant rectangular segments along the vertical axis. The center of the image aligns with the midpoint of all these vertical axis bins. The width of each bin can either be preassigned as a constant value or determined based on the specific characteristics of the image. The fixed width is set to 2.5 cm and the adaptive version is the median distance divided by 20, the same as that described in the annuli design \cite{liu2020automatic}.
    \end{enumerate}

 Features that consolidate data globally, aggregating the characteristics of all individual bloodstains within a single spatter pattern, would be termed ``global features." On the other hand, features that consolidate data within segregated regions, such as annuli or rectangular bins, would be referred to as ``local features". 
 
 Next, we describe a few methods to summarize the ellipse features. We consider a dataset $\{x_i\}^n_{i=1}$, where $x_i$ can be a scalar, a vector (depending on the specific feature of bloodstain) or the index of a local region; and $n$ is the number of ellipses in a pattern or the counts of separated regions. 
    
    \begin{enumerate}
 
        \item Mean: The average of a specific stain feature,  $\Bar{x} = \frac{1}{n}\sum^n_{i=1} x_i$.
        \item Standard deviation (SD): The standard deviation of a specific stain feature$$\big( \frac{1}{n}(\sum^n_{i=1} x_i - \Bar{x})^2 \big)^{\frac{1}{2}}.$$  
        \item Counts: The number of stains that meet a specific condition $\mathbf{C}$, $m = \sum^n_{i=1} I(x_i\in \mathbf{C})$.
        \item Ratio: The proportion of stains that a specific condition $\mathbf{C}$, $r = m/n$.
        \item Index: Subjects that satisfy a specific condition within the dataset, that is, $k = \{ i \vert x_i\in \mathbf{C}, i = 1,...,n\}$.
    \end{enumerate}

Importantly, for angular stain features like \textit{Orientation} and \textit{Impact angle}, standard methods such as calculating the mean and variance may not be appropriate. For instance, if we apply the conventional methods designed for variables in the Euclidean space, an average angle of $180\degree$ for $10\degree$ and $350\degree$. However, the true average angle should be $0\degree$ because those angles belong to a circular space. As a result, we use circular statistics to compute the mean and variance of angular features following the methods in \cite{zou2022towards}:

    \begin{enumerate}
    \setcounter{enumi}{5}
        \item Angular variance: 
For an angular variable $\alpha$, we map it into a 2-D space with polar coordinates:

        $$\alpha \in \lbrack0,360) \mapsto \textbf{u} = (\cos \alpha, \sin \alpha) \in \mathbb{R}^2.$$
        
        Then the average unit vector is 

        $$\Bar{\textbf{u}} = \frac{1}{n}\sum^n_{i =1}\textbf{u}_i = (\frac{1}{n}\sum^n_{i=1} \cos \alpha_i, \frac{1}{n}\sum^n_{i=1} \sin \alpha_i).$$
        
        The variance of $\alpha$ is then defined as the difference between one and the length of average unit vector:

        $$\text{Var}(\alpha) \coloneqq 1 - \big \{ (\frac{1}{n}\sum^n_{i=1} \cos \alpha_i)^2 + (\frac{1}{n}\sum^n_{i=1} \sin \alpha_i)^2 \big \}^{\frac{1}{2}}.$$

        \item Spherical features: As noted in \cite{zou2022towards}, the incident direction is inherent in a 3-D view. This suggests that the relationship between the orientation angle and the impact angle should be explored. To effectively combine the information from these two angular variables, we define the unit vector of the incident direction for each stain with impact angle $\alpha$ and orientation angle $\beta$ as:

        \[
        \mathbf{m} = (-\cos\alpha\cos\beta, -\cos\alpha\sin\beta, \sin\alpha).
        \]
        
        Subsequently, the scatter matrix $T$, given by:
        
        \[
        T = \sum^{n}_{i=1}\mathbf{m}_i\mathbf{m}^T_i,
        \]
        
        for all the stains with $i$ ranging from 1 to $n$, is interpreted as the inertia tensor of a rigid body about its origin. This body is composed of equal weight particles positioned at each location of $\mathbf{m}_i$. \cite{zou2022towards} have demonstrated that the eigenvalues of $T$, denoted as $t^1$, $t^2$, $t^3$ ($t^1 \geq t^2 \geq t^3$), and the corresponding eigenvectors $\mathbf{t}^1$, $\mathbf{t}^2$, $\mathbf{t}^3$, provide valuable insights into the shape of $T$. The ratio of $t^1$ to $t^2$ represents the symmetry of the distribution of the vector $\mathbf{m}_i$, while the determinant $\det(T)$ can be used to summarize the dispersion.

    \end{enumerate}

\section{Statistical Analysis}\label{sec4}
\subsection{Exploratory Data Analysis}

After extracting numerical features from the images and excluding patterns unable to detect at least 2 bloodstains, we obtain a dataset of 114 independent patterns (65 gunshot patterns and 49 impact patterns), each consisting of 48 features. The complete feature list, containing the names and descriptions, is given in Table \ref{table: Feature table}.

Next, we conduct an exploratory analysis. In particular, we focus on five features \textit{mean\_maj\_len}, \textit{mean\_min\_len}, \textit{mean\_area}, \textit{mean\_solidity}, and \textit{fract1\_ring\_15\_25}, and study how they differ between gunshot and impact patterns. Using boxplots for each feature, we examine their central tendencies, dispersions, and presences of outliers to discern their distributional characteristics across different blood pattern generating mechanisms.

 \begin{figure}
        \includegraphics[width = \textwidth]{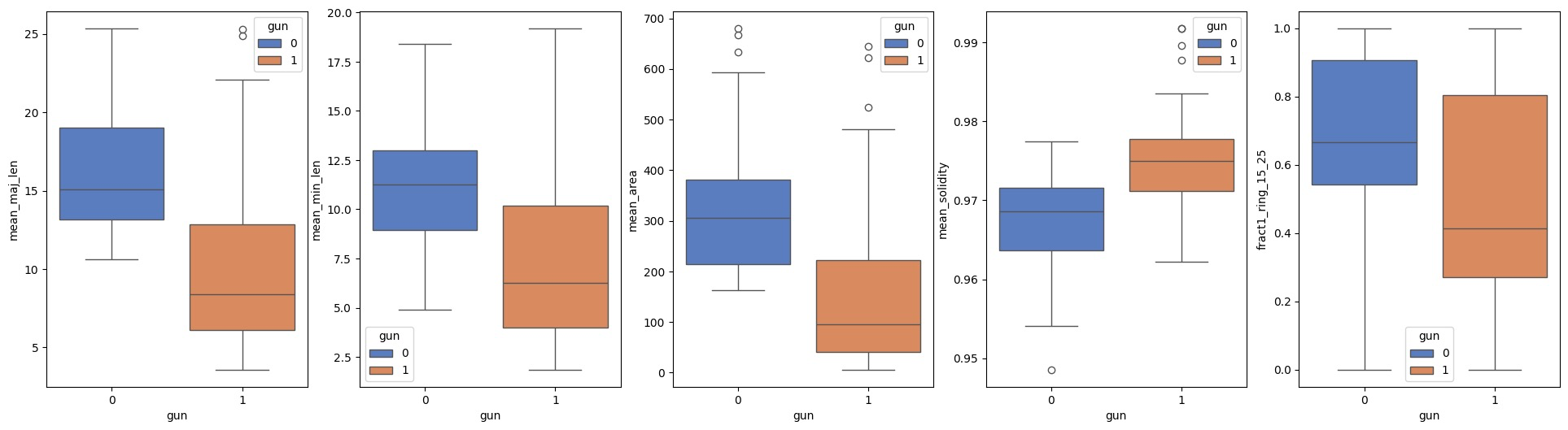}\caption{Boxplot of Features Group by Label \textit{gun}, for feature \textit{mean\_maj\_len}, \textit{mean\_min\_len}, \textit{mean\_area}, \textit{mean\_solidity}, and \textit{fract1\_ring\_15\_25}, where \textit{gun = 1} stands for gun-shot patterns.}
        \label{figure:boxplot}
    \end{figure}

From the boxplots in Figure \ref{figure:boxplot}, we find that features such as \textit{mean\_maj\_len}, \textit{mean\_min\_len}, \textit{mean\_area}, and \textit{mean\_solidity} demonstrate notable discrepancies between the two classes, suggesting their strong discriminative potential in the classification task. More specifically, gunshot patterns exhibit lower major and minor length values, lower mean area values, and higher mean solidity values compared to impact patterns. Gunshot patterns also have a higher level of variability compared to impact patterns in general except for \textit{mean\_solidity}. For \textit{fract1\_ring\_15\_25}, the mean/median difference is more subtle, indicating a less pronounced distinction between the classes based on this feature alone. 

The variability in distribution, as depicted by the interquartile ranges (IQR) and the presence of outliers, provides further depth to our analysis. Features with smaller IQRs, such as \textit{mean\_area} and \textit{mean\_solidity}, imply a narrower diversity in the data associated with the generating mechanism. This contrasts with features like \textit{fract1\_ring\_15\_25} with a wider IQR. These observations highlight the importance of considering multifaceted classification models. While some features exhibit clear class separation, others may contribute to model robustness by providing nuanced insights into the data's underlying structure. Therefore, it is necessary to explore various classification models that can fully utilize the information from the data.

\begin{table}[H]

\centering

    \begin{tabularx}{380pt}{|l X|}
\hline
 Feature name & Description  \\ [0.5ex] 
 \hline
        num\_stains & The quantity of bloodstains within a or pattern \\
        mean\_maj\_length & Mean major axis length of the bloodstains \\
        mean\_min\_length & Mean minor axis length of the bloodstains \\
        mean\_area & Mean area of the bloodstains \\ 
        mean/sd\_ratio\_dis & Mean and SD of normalized distance of the bloodstains \\ 
        sd\_epslion & SD of epsilons of the bloodstains \\
        sd\_impact\_angle & SD of impact angles of the bloodstains \\ 
        mean/sd\_solidity & Mean and SD of solidity of the bloodstains \\
        num\_large\_1/75 & Counts of bloodstains larger than $(0.1 \text{ or } 0.075mm)^2\pi$ \\
        ratio\_large\_1/75 & Ratio of bloodstains larger than $(0.1 \text{ or } 0.075mm)^2\pi$ \\
        fract1/75\_ring\_j\_j+10 & Ratio of bloodstains larger than $(0.1 \text{ or } 0.075mm)^2\pi$ in fixed rings from j to j+10, j = 5, 15, 25 \\ 
        adp\_fract1/75\_ring\_15\_25/25\_31 & Ratio of bloodstains larger than $(0.1 \text{ or } 0.075mm)^2\pi$ in adaptive rings from 15 to 25/ 25 to 31 \\
        num1/75\_rec\_j\_j+10 & Counts of bloodstains larger than $(0.1 \text{ or } 0.075mm)^2\pi$, in fixed rings from j to j+10, j = 5,15,25 \\ 
        fract1/75\_rec\_j\_j+10 & Ratio of bloodstains larger than $(0.1 \text{ or } 0.075mm)^2\pi$, in fixed rings from j to j+10, j = 5,15,25 \\
        (adp\_/rec\_)i & The index of the (adaptive) ring/rectangular bin that contains the most of the stains out of all the bins \\
        (adp\_/rec\_)m & The maximum number of stains found within a single (adaptive) ring/rectangular bin \\
        rec(\_adp)\_bin\_ratio & The ratio of (adp\_)i to rec\_i  \\
        spheri\_ratio &  The ratio between the second and the third eigenvalue of scatter matrix \\
        spheri\_det & The determinant of scatter matrix \\ 
\hline

    \end{tabularx}
    \caption{Features and their descriptions ( ``/" refers to  ``or", and the content inside ``()" can be omitted. For instance, item \textit{num\_large\_1/75} in the table refers to feature \textit{num\_large\_1} and \textit{num\_large\_75}; item \textit{(adp\_/rec\_)i} contains feature \textit{i}, \textit{adp\_i} and feature \textit{rec\_i}).}
    \label{table: Feature table}
\end{table}

\subsection{Models and Results}\label{sec: models}
We apply several machine learning algorithms to classify gunshot backspatters and impact spatters. Among them, XGBoost and random forest achieve the highest classification accuracy. Given that the feature importance in a boosting or bagging model is greatly influenced by the random starting point, and that the value of the importance from each fitting round is ordinal data, summing these importance to calculate an average feature importance is not appropriate. Consequently, we introduce the \textit{Stability Importance Score} (SIS), which represents the frequency with which a feature ranks among the top 10 important features across all fittings. This approach focuses solely on the rank of the feature, making it more interpretable and aligning well with the definition of feature importance. In the following, we provide a detailed discussion of the results. 

\subsubsection{XGBoost}
XGBoost, short for ``Extreme Gradient Boosting", is a widely used ensemble learning method, known for its exceptional performance in both classification and regression tasks, especially on large datasets \cite{chen2016xgboost}. Notably, XGBoost excels at handling missing data without requiring extensive pre-processing and offers built-in support for parallel processing, ensuring fast model training on sizable datasets. 

We split the dataset into a 75\% training dataset and a 25\% test dataset in each round of fitting. After conducting hyperparameter tuning through grid search, the average overall accuracy in 20,000 fittings is 92. 89\%. When considering the blood-to-target (BT) distance as recorded in \cite{attinger2019data} and \cite{attinger2018data}, the accuracy in one specific fitting, reaches 100\% for BT distances less than 60 cm, 96\% for BT distances less than 120 cm, and up to 100\% for patterns with a BT distance greater than 120 cm. This results in a total accuracy of 96.55\%.
Over 2000 fittings, the average accuracy is 93.39\%, 95.44\% and 91.84\%, respectively for BT distances within 30 cm, 60 cm, and 120 cm.

\begin{figure}[H]
\includegraphics[width = \textwidth]{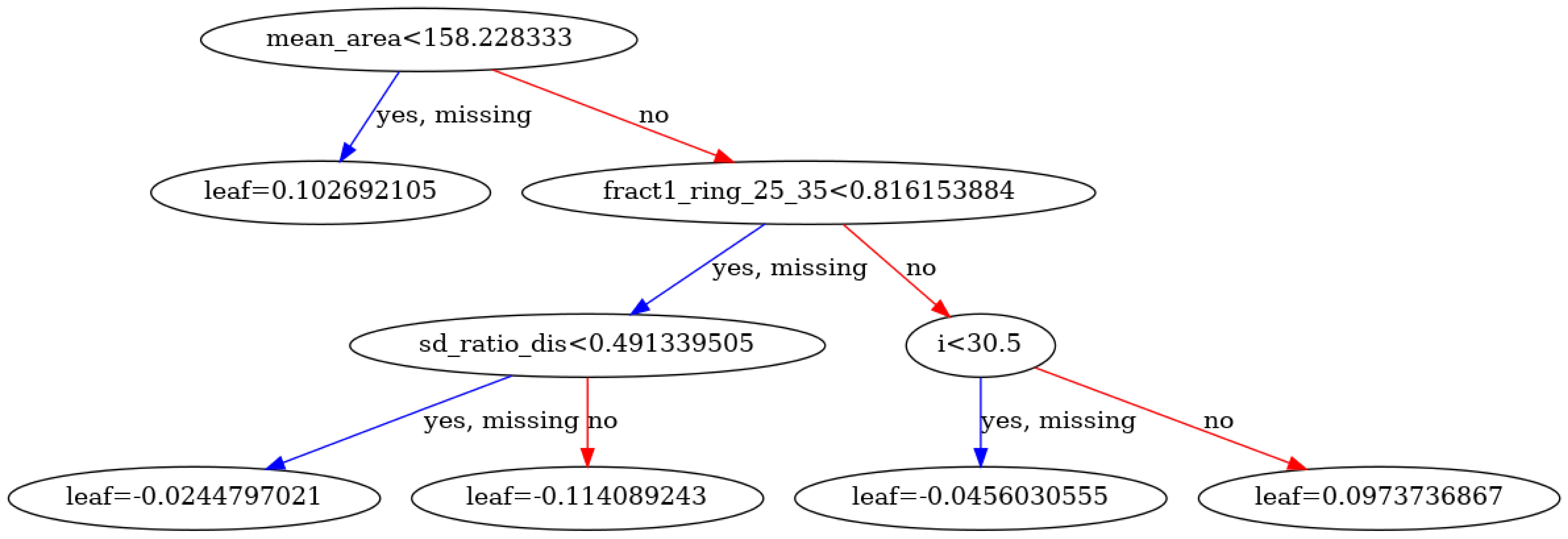}\caption{An Example of the Fitted Tree Plot from the XGBoost Model.}
\label{figure:XGBoost_tree_plot}
\end{figure}

As an illustration, we show an example of the XGBoost tree in Figure \ref{figure:XGBoost_tree_plot}. For each node with ``leaf" value, it represents the raw score for being in ``gun group." If there is a leaf with value $x$, after being converted to a probability score by applying logistic transformation, the probability will be calculated by $(1+\exp\{-x\})^{-1}$. For instance, the first leaf on the left-hand side indicates a probability score of $(1+\exp \{(-1)\times (-0.02448)\})^{-1} = 0.494$; the node ``$i < 30.5$" indicates that if the index of the ring that contains the most of the stains among all rings in the pattern is less than or equal to 30, then the raw score for the pattern would be approximately $-0.0456$, and $0.0974$ otherwise.

\begin{figure}[H]
\includegraphics[width = \textwidth]{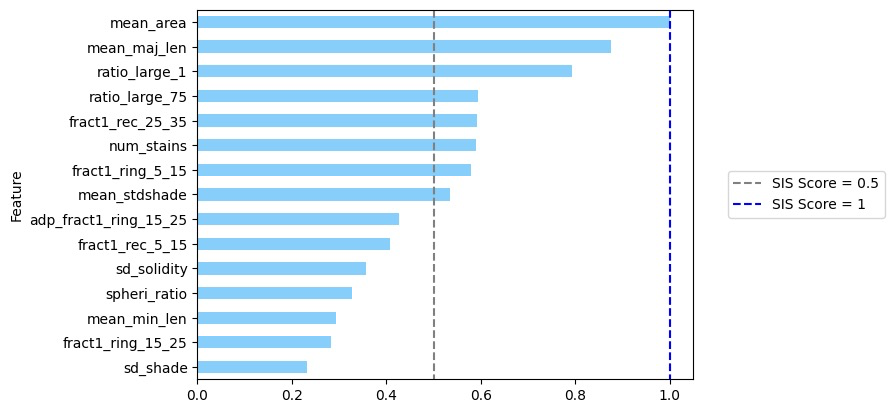}\caption{SIS scores for XGBoost model: Frequency of each feature when selected among the Top 10 important features over 2000 replications.}
\label{figure:XGBoost_importance}
\end{figure}

We conduct model fitting 2000 times and record the SIS score, i.e., the frequency of each feature when it is selected among the top 10 important features for the model in a single fitting. The features with the highest SIS scores are shown in Figure \ref{figure:XGBoost_importance}. We find that feature \textit{mean\_area}, which stands for the average size of the stains, has the highest SIS score of being among the most important features for discriminating between gun patterns and impact patterns. This finding is consistent with previous studies \cite{liu2020automatic}. The inclusion of pattern regions appears necessary, as evidenced by 6 local features such as \textit{fract1\_rec\_25\_35} and  \textit{adp\_fract1\_ring\_15\_25}, having high frequencies over 0.5. There are seven features ranked in the top 10 important features for more than half of the total fitting rounds. As features that have never been used in previous studies to build classifiers, shade-related features show great potential to help identify patterns. Furthermore, the presence of the feature \textit{spheri\_ratio} in this figure highlights the importance of including directional statistics to build an efficient boosting model.

\subsubsection{Random Forest} 
Random forest is another popular ensemble method used for classification, regression, and other tasks \cite{breiman2001random}. It constructs multiple decision trees during training and the final output in classification is determined by majority vote.  Unlike boosting, random forest does not automatically handle missing values. In our data analysis, we encounter missing values in certain situations, such as instances where the ``ratio" in specific regions of a pattern appears as \textit{NaN} when the population in those regions is zero. To address this, we implement $k$-Nearest Neighbor ($k$-NN) imputation and zero imputation to fill in the missing values before proceeding with model fitting and hyperparameter tuning. 

$k$-NN imputation is a method that aims to find the $k$ nearest neighbors for a missing data point. It works by identifying complete instances (those without missing values) and then replaces the missing data with mean of the neighbors for numerical features. Following a grid search for optimal hyperparameters, we choose $k=10$ and the average accuracy over 2000 loops is 89.91\%, slightly less than that of XGBoost. Taking into account the distance of the BT, as shown in Table \ref{tab:accuracy_comparison}, it correctly predicts 91.92\% of labels within 30 cm and 92.27\% within 60 cm. For patterns with greater BT distances, the accuracy decreases to 88.37\%. Overall speaking, the performance of the random forest model is slightly worse than that of the XGBoost. 
    
\begin{figure}[H]
    \includegraphics[width = \textwidth]{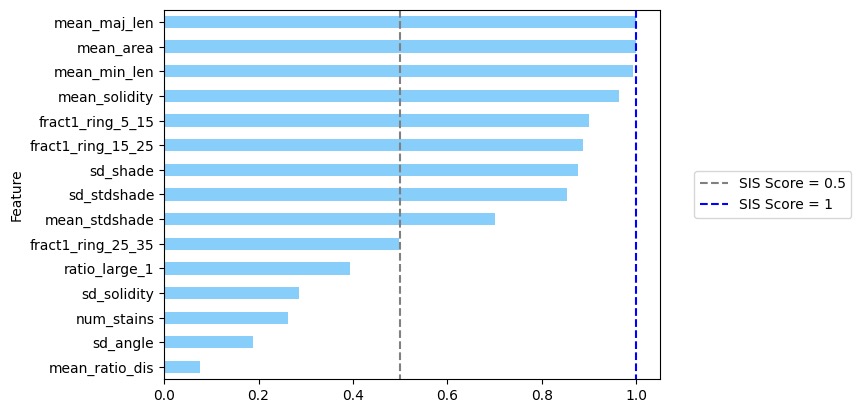}\caption{SIS scores for random forest model: Ranking of the frequency of the top 10 important features in 2000 random forest model fits with $k$-NN imputation.}
    \label{figure:RandomForest_KNN_importance}
    \end{figure}

In Figure \ref{figure:RandomForest_KNN_importance}, we present the frequencies of features that appear in the top 10 important features across 2000 rounds of modeling using the random forest classifier. The important features show a more stable tendency, with 9 features having a SIS score of over 0.5, and 2 of them consistently maintaining their position within the top 10 features throughout all 2000 rounds. In comparison, XGBoost exhibits 8 features with a SIS score over one-half, and only 1 consistently remains in the top 10. We also notice that local features play crucial roles in decision making, with 3 of them listed in the 15 most frequent features.
    
While $k$-NN imputation might be a better choice for missing values in general, in our dataset where missing values exist in the features \textit{fract1\_ring\_15\_25}, \textit{fract75\_ring}-\textit{\_15\_25}, \textit{fract1\_ring\_5\_15}, \textit{fract75\_ring\_5\_15}, which are all local features consolidating by the ratio method. Missing values are generated when there is no single stain in the specific region. As a result, both the count of stains that satisfies certain conditions and the total counts of all stains in the region would be 0, and this would lead to a ``0/0", producing \textit{NaN} in the feature dataset. Zero imputation makes sense as it can reveal that the information that none of the stains in the region meets our requirements, and that this region contains no stains at all, has already been passed to the model by the counting-consolidating local features.

We adopt zero imputation and the classifier achieves an accuracy of 90.27\% based on 2000 random iterations, showing a similar performance to that model trained on the $k$-NN-imputed dataset, and also falls short of XGBoost. Taking into account the BT distance, the accuracy is 93.83\% within 30 cm, 92. 63\% within 60 cm, and 90. 27\% overall.

\begin{figure}[H]
    \includegraphics[width = \textwidth]{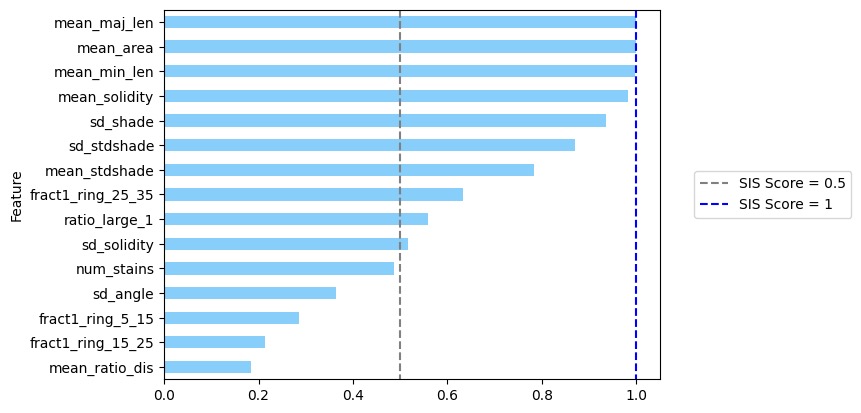}\caption{SIS scores for random forest model: Ranking of the frequency of the Top 10 important features in 2000 fits with zero imputation.}
    \label{figure:RandomForest_zero_importance}
    \end{figure}

In Figure \ref{figure:RandomForest_zero_importance}, we observe an even more stable set of essential features, with 10 of them having a SIS score above 0.5 to be ranked among the top 10 important features, and 3 of them consistently maintaining their position in every fitting. The fact that the four size-related features consolidate as the top four features with a SIS score of one aligns with the results obtained from random forest classifiers trained on the $k$-NN-imputed dataset. However, it should be noted that shade-related features exhibit higher frequencies compared to previous models, especially compared to the XGBoost models. Considering the slight advantage in accuracy and consistency of important features, zero imputation appears to be a better alternative for random forest classifiers. 

Compared to the random forest model previously developed by \cite{liu2020automatic}, which achieved an average accuracy of 86. 10\% for all distances from BT, our method with $k$-NN imputation offers comparable accuracy for all distances from BT. Furthermore, our RF classifier with zero imputation not only achieves a significantly higher classification accuracy of 90.26\%, but also benefits from a more interpretable and concise dataset.

\begin{table}[ht]
    \centering
    \caption{Accuracy Comparison for Different BT Distances}
    \begin{tabularx}{318pt}{|m{3cm}|m{1.3cm}|m{1.5cm}|m{1.5cm}|m{1.7cm}|}
    \hline
        \textbf{Model} & For all $d$ & $d \leq 30 cm$ & $d \leq 60 cm$ & $d \leq 120 cm$ \\
    \hline
        XGBoost \rule{0pt}{3ex}  & 0.9289 & 0.9339 & 0.9544 & 0.9184 \\
    \hline
        Random Forest \newline ($k$-NN imputation) & 0.8992 & 0.9192 & 0.9227 & 0.8837 \\
    \hline
        Random Forest \newline (zero imputation) & 0.9027 & 0.9383 & 0.9263 & 0.8854 \\
    \hline
    \end{tabularx}
    \label{tab:accuracy_comparison}
\end{table}

\subsubsection{Other Classifiers}
We have also explored other commonly used classification methods in our analysis. For example, we choose to use Quadratic Discriminant Analysis (QDA) \cite{hastie2009elements} over Linear Discriminant Analysis (LDA), which has been previously used in BPA  \cite{arthur2018automated}, for the reason that LDA assumes equal variances between two classes, which is too restrictive.  The training accuracy of QDA in our analysis is around 40-65\%, which falls behind that of XGBoost and Random Forest. We also consider the support vector machine \cite{boser1992training} with a Gaussian kernel and the accuracy in training is between 60 and 70\%, better than the QDA method but still behind the boosting and bagging models.




\section{Discussion}
\label{sec: discussion}
In this study, we develop a comprehensive method to classify gunshot backspatter patterns from impact beating spatter patterns, including image pre-processing, identification of individual stain features, determination of how these features should be consolidated, where the consolidation methods are applied, and model fitting for classification. The impressive classification accuracy indicates the great potential statistical machine learning methods applied to bloodstain pattern analysis, and further to other forensic problems to get useful insights during the investigation of violent crimes. Especially, the XGBoost models, which achieve an impressive average accuracy of up to 92.89\%, surpass all previous models developed on the same set of images, even those with more features, which shows the huge potential of applying machine learning methods to quantitative analysis in forensic science. In addition to the classifier based on the boosting model, we design an improved method for measuring feature importance, providing reliable insights into the characteristics of bloodstains despite the randomness of each single model.

A core problem in bloodstain pattern analysis is how to transform an image into numerical data. In our study, we preprocessed the images for ellipse detection and utilized the inherent elliptical properties for feature engineering. In addition to shape features that have been widely explored in other studies, we took the shade of the stains into consideration, and the results of the models proved this idea to be helpful in building effective classifiers. In addition to darkness, although stains are generally of red color, their tone may vary. Considering their RGB values in further studies can extend the feature space by three more dimensions.

In addition to the dataset we studied in this paper, we also attempted to extend our analysis to other datasets from previous studies such as \cite{arthur2018automated} and \cite{laan2015bloodstain}. For example, from \cite{arthur2018automated}, we collected two impact patterns provided as TIFF-format images, with a resolution of approximately 6.4 pixels per millimeter. The resolution information for the impact patterns from \cite{laan2015bloodstain} was unavailable. We then pre-processed and analyzed the datasets following the same pipeline as described in this paper. Interestingly, all classifiers that we have considered yielded a much lower classification accuracy around 50\% for both datasets. One possible explanation is the inherent variability in the datasets collected from different sources. Bloodstain patterns can vary due to factors such as the experimental setup for generation, lighting conditions, camera equipment, shooting angles, and even the specific context in which the data was collected. As a result, a model trained on one data set generated and collected in the same environmental settings may not effectively capture the full range of possible patterns, leading to unsatisfactory performance when applied to diverse data sources. Improving data quality (e.g. better image resolution) and developing flexible models that can work well for diverse data sources are promising future work directions.



\printbibliography

\end{document}